\documentclass[runningheads]{llncs}

\usepackage[mobile]{eccv}

\usepackage{eccvabbrv}

\usepackage{graphicx}
\usepackage{booktabs}
\usepackage{multirow}
\usepackage{makecell}
\usepackage{amssymb}
\usepackage{geometry} 
\usepackage{array}

\usepackage[pagebackref,breaklinks,colorlinks,citecolor=eccvblue]{hyperref}
\usepackage{orcidlink}

\usepackage{graphicx}

\newcommand{\hindiArvind}{\raisebox{-0.15ex}{\includegraphics[height=0.8em]{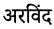}}}
\newcommand{\hindiAravid}{\raisebox{-0.15ex}{\includegraphics[height=0.8em]{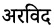}}}

\newcommand{\thaiBiggest}{\raisebox{-0.15ex}{\includegraphics[height=0.8em]{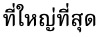}}}
\newcommand{\thaiBiggestWrong}{\raisebox{-0.15ex}{\includegraphics[height=0.8em]{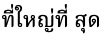}}}

\newcommand{\rusA}{\raisebox{-0.15ex}{\includegraphics[height=0.8em]{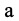}}}

\newcommand{\rusH}{\raisebox{-0.15ex}{\includegraphics[height=0.8em]{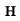}}}

\newcommand{\rusO}{\raisebox{-0.15ex}{\includegraphics[height=0.8em]{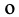}}}

\newcommand{\rusAHO}{\raisebox{-0.15ex}{\includegraphics[height=0.8em]{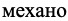}}}

\begin{document}

\title{MDPBench: A Benchmark for Multilingual Document Parsing in Real-World Scenarios} 

\titlerunning{MDPBench}

% Second Author\inst{2,3}\orcidlink{1111-2222-3333-4444} \and
% Third Author\inst{3}\orcidlink{2222--3333-4444-5555}}
\author{Zhang Li\inst{1}\thanks{Core contribution.}, Zhibo Lin\inst{1}$^\star$, Qiang Liu\inst{2}, Ziyang Zhang\inst{1}, Shuo Zhang\inst{1},\\Zidun Guo\inst{1}, Jiajun Song\inst{1}, Jiarui Zhang\inst{2}, Xiang Bai\inst{1}, Yuliang Liu\inst{1}}
\authorrunning{Li.~et al.}

% Springer Heidelberg, Tiergartenstr.~17, 69121 Heidelberg, Germany

\institute{\textsuperscript{1}Huazhong University of Science and Technology, \textsuperscript{2}Kingsoft Office}

\maketitle

\begin{abstract}
  We introduce \textbf{M}ultilingual \textbf{D}ocument \textbf{P}arsing \textbf{Bench}mark, the first benchmark for multilingual digital and photographed document parsing. Document parsing has made remarkable strides, yet almost exclusively on clean, digital, well-formatted pages in a handful of dominant languages. No systematic benchmark exists to evaluate how models perform on digital and photographed documents across diverse scripts and low-resource languages. MDPBench comprises 3,400 document images spanning 17 languages, diverse scripts, and varied photographic conditions, with high-quality annotations produced through a rigorous pipeline of expert model labeling, manual correction, and human verification. To ensure fair comparison and prevent data leakage, we maintain separate public and private evaluation splits. Our comprehensive evaluation of both open-source and closed-source models uncovers a striking finding: while closed-source models (notably Gemini3-Pro) prove relatively robust, open-source alternatives suffer dramatic performance collapse, particularly on non-Latin scripts and real-world photographed documents, with an average drop of 17.8\% on photographed documents and 14.0\% on non-Latin scripts. These results reveal significant performance imbalances across languages and conditions, and point to concrete directions for building more inclusive, deployment-ready parsing systems. Source available at \url{https://github.com/Yuliang-Liu/MultimodalOCR}.

\end{abstract}

\begin{figure}
    \centering
    \includegraphics[width=1\textwidth]{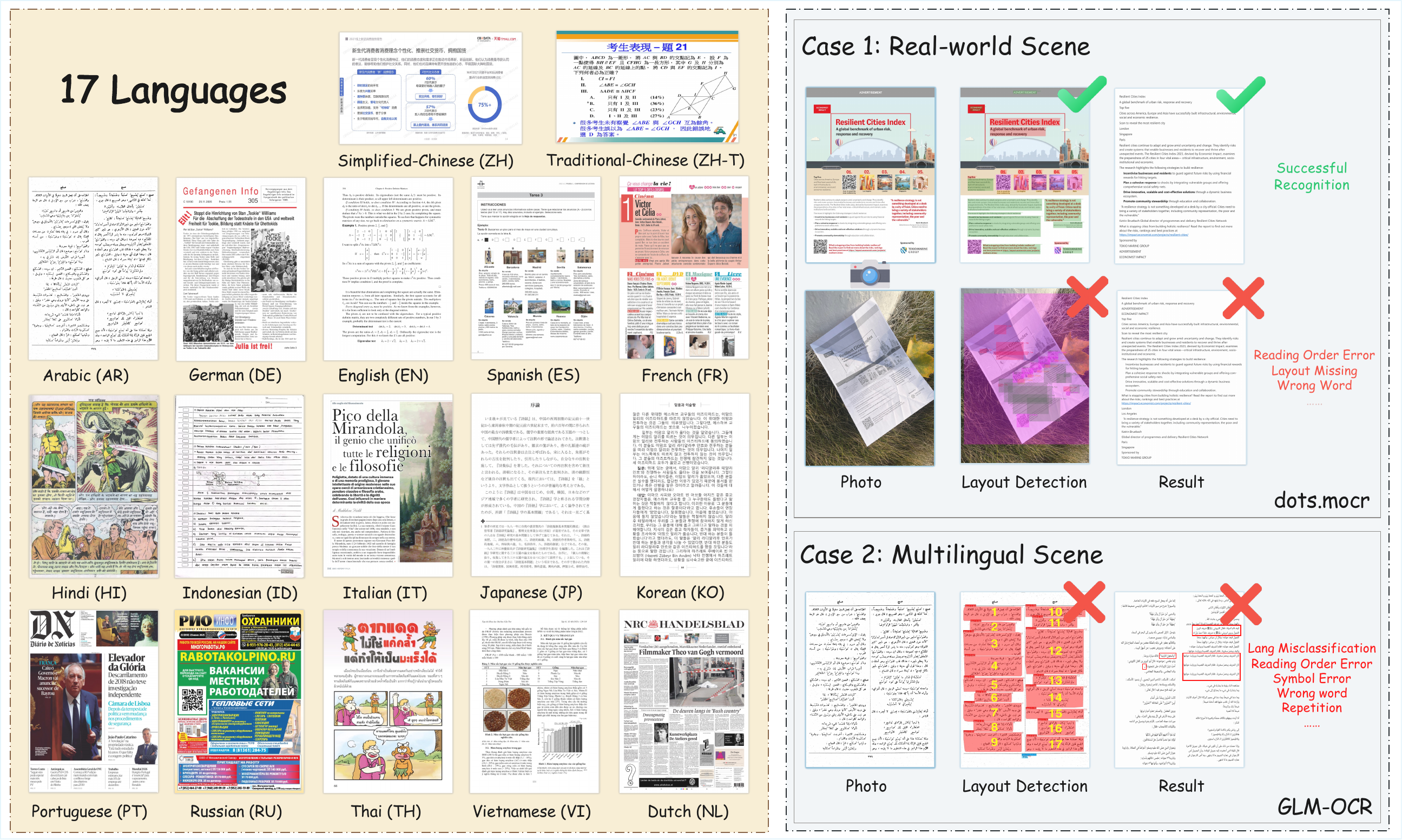}
    \caption{Overview of the MDPBench.}
    \label{fig:overview}
\end{figure}

\section{Introduction}
\label{sec:intro}

Document parsing bridges the gap between visual information and machine-readable text by converting document images into structured, serialized representations. As a cornerstone for building high-quality pre-training corpora for LLMs, it directly influences the scale and fidelity with which human knowledge is transferred into machine intelligence. Significant progress has been made in document parsing, with numerous methods~\cite{mineru, ppstruct3, monkeyocr, monkeyocr1.5, dotsocr, paddleocr-vl, paddleocr-vl-1.5, olmocr, olmocr2} continuously pushing state-of-the-art performance on benchmarks such as OmniDocBench~\cite{omnidocbench} and olmOCR-Bench~\cite{olmocr}. However, these benchmarks predominantly focus on digital-born and scanned documents in limited languages. This limitation leads existing document parsing models to exhibit a certain bias toward inputs from standardized, high-resource languages, with performance often declining in multilingual and challenging photographed document scenarios.

Multilingual and photographed document parsing plays a crucial role in the development of general-purpose AI systems. On the one hand, multilingual documents encapsulate diverse knowledge from different regions and cultural contexts. Enhancing multilingual parsing capabilities enables more comprehensive access to global knowledge, helping mitigate potential biases arising from imbalanced language distributions. On the other hand, a substantial portion of real-world documents exists only in photographed form, including historical archives, paper receipts, books, and handwritten notes, often without corresponding digital versions. Effectively parsing such documents is essential for unlocking the value of large-scale unstructured data, thereby supporting more scalable and higher-quality LLM pre-training.

To promote research on document parsing in multilingual and photographed document scenarios, we introduce Multilingual Document Parsing Benchmark (MDPBench), which consists of 3,400 document images across 17 languages, as shown in Fig.~\ref{fig:overview}. We first collected electronic documents from publicly accessible websites, covering a wide range of document types, including academic papers, business reports, handwritten notes, newspapers, textbooks, and comics from different countries. To ensure annotation quality, we adopt a multi-stage annotation pipeline that combines automatic labeling and cross-verification by multiple expert models, followed by manual correction and human verification. To obtain photographed documents that better reflect real-world conditions, we printed the collected electronic documents or displayed them on screens, and captured them under diverse environments and conditions, including indoor and outdoor scenes, physical deformation, image degradation, varying camera orientations, and background variations.
To prevent data leakage and targeted training, the dataset is divided into 2,720 publicly available benchmark samples and 680 private evaluation samples. Researchers can submit their models to the official evaluation website for assessment on the private benchmark.

\begin{table*}[t]
\centering
\caption{Comparison of document parsing benchmarks.\textbf{DB}: Digital-Born; \textbf{Photo.}: Photographed; \textbf{SR}: Screen Re-photograph; \textbf{PD}: Physical Deformation; \textbf{ID}: Image Degradation; \textbf{CO}: Camera Orientations; \textbf{BV}: Background Variation.}
\label{tab:benchmark_final_pro}
\small
\setlength{\tabcolsep}{3.5pt}
\renewcommand{\arraystretch}{1.2}
\resizebox{\linewidth}{!}{
\begin{tabular}{c *{8}{c} c c *{5}{>{\centering\arraybackslash}p{0.65cm}}}
\toprule
\multirow{2}{*}{\textbf{Benchmark}} & \multicolumn{8}{c}{\textbf{Languages}} & \multirow{2}{*}{\textbf{Type}} & \multirow{2}{*}{\makecell[c]{\textbf{Image}\\\textbf{Count}}} & \multicolumn{5}{c}{\textbf{Photograph Conditions}} \\
\cmidrule(lr){2-9} \cmidrule(lr){12-16}
& Num & ZH & EN & FR & ES & RU & AR & Others & & & SR & PD & ID & CO & BV \\
\midrule
FoxPage~\cite{foxpage} & 2 & \checkmark & \checkmark & & & & & --- & DB & 212 & & & & & \\
olmOCR-Bench~\cite{olmocr} & 1 & & \checkmark & & & & & --- & DB & 1402 & & & & & \\
OmniDocBench-v1.5~\cite{omnidocbench} & 2 & \checkmark & \checkmark & & & & & --- & DB & 1355 & & & & & \\
DocPTBench~\cite{docptbench} & 2 & \checkmark & \checkmark & & & & & --- & DB / Photo. & 2362 & & \checkmark & \checkmark & \checkmark & \checkmark \\
Real5-OmniDocBench~\cite{real5-omnidocbench} & 2 & \checkmark & \checkmark & & & & & --- & DB / Photo. & 6775 & \checkmark & \checkmark & \checkmark & & \\
\midrule
\textbf{MDPBench} & 17 & \checkmark & \checkmark & \checkmark & \checkmark & \checkmark & \checkmark & \makecell[c]{DE, HI, ID, IT, \\ JP, KO, NL, PT, \\ TH, VI, ZH-T} & DB / Photo. & 3400 & \checkmark & \checkmark & \checkmark & \checkmark & \checkmark \\
\bottomrule
\end{tabular}
}
\vspace{1ex}
\raggedright
\end{table*}

We conduct a comprehensive evaluation of general-purpose vision-language models (VLMs), specialized VLMs, and pipeline-based systems on MDPBench. The results reveal that: (1) open-source models still lag behind proprietary models, with Gemini-3-Pro~\cite{gemini3pro} outperforming the strongest open-source model, dots.mocr~\cite{dotsmocr}, by 7.9\% in photographed scenarios; (2) all methods experience notable performance degradation on photographed documents, with an average drop of 17.8\%; and (3) performance on non-Latin-script languages is consistently lower, showing an average decrease of 14.0\%. Overall, MDPBench highlights the limitations of current document parsing approaches and provides a standardized benchmark for evaluating multilingual text understanding and OCR capabilities in general VLMs.

We summarize the contributions as follows: 

\begin{itemize}
    \item We introduce \textbf{MDPBench}, the first benchmark specifically designed for multilingual digital and photographed document parsing. It comprises 3,400 document images spanning 17 languages and diverse document types, with high-quality annotations obtained through a rigorous multi-stage pipeline, including expert model labeling, manual correction, and human verification.

    \item  A comprehensive evaluation shows that open-source models still lag behind the state-of-the-art proprietary model, Gemini-3-Pro~\cite{gemini3pro}. Moreover, all models exhibit significant performance degradation on photographed documents and non-Latin scripts, highlighting the limitations of current approaches in real-world multilingual scenarios.
\end{itemize}

\begin{figure}[tbp]
    \centering
    \includegraphics[width=0.9\textwidth]{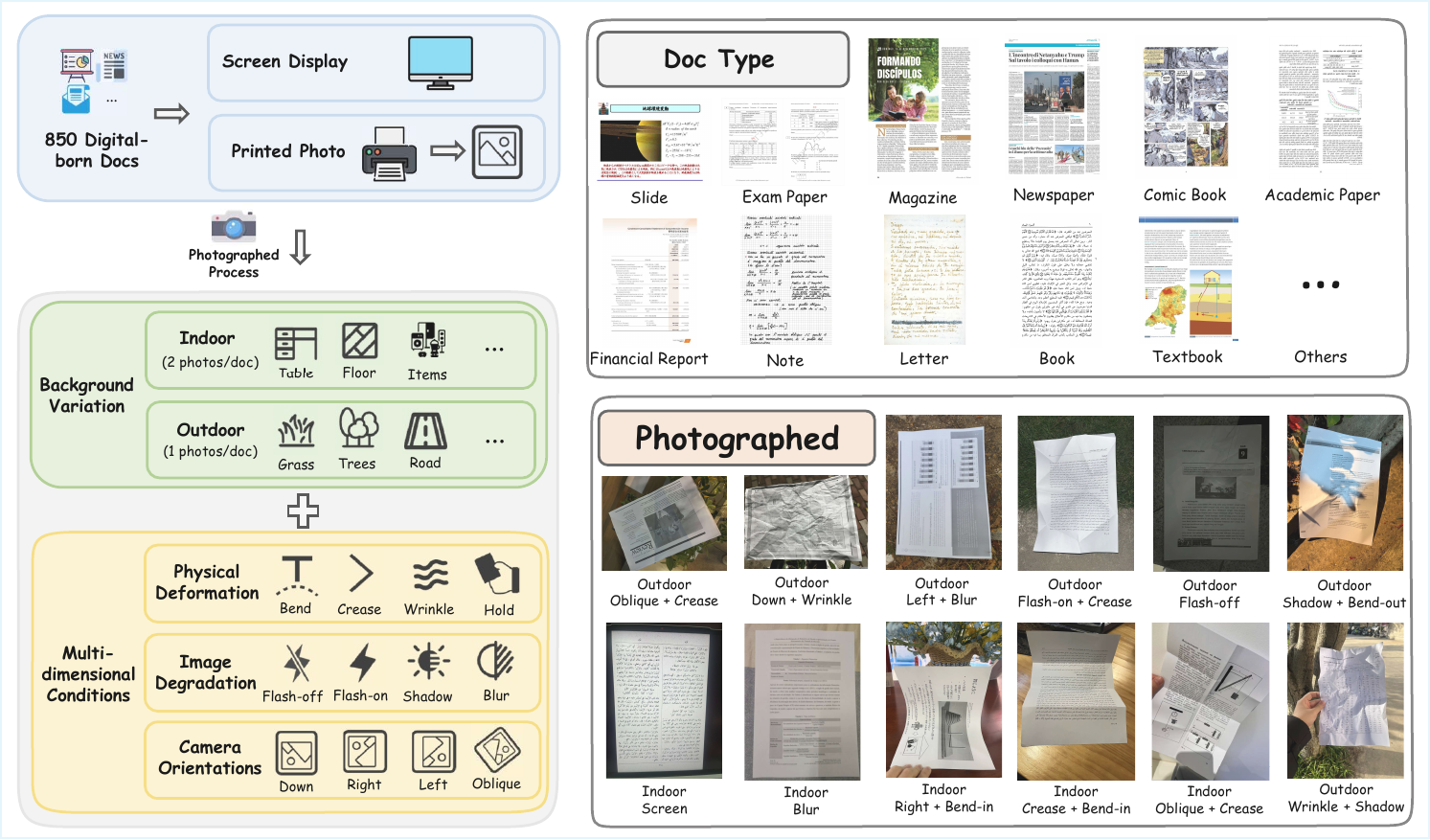}
    \caption{Visualization of digital-born and photographed document images.}
    \label{fig:photo}
\end{figure}

\section{Related Work}

\subsection{Document Parsing Methods}

Document parsing methods can be broadly categorized into traditional pipeline methods, general vision–language models (VLMs), end-to-end specialized VLMs, and multi-stage specialized VLMs. Traditional pipeline~\cite{mineru,ppstruct3,marker,docling} methods typically follow a fixed workflow: they first perform layout detection, then detect and recognize elements, merge the recognized elements, and finally reconstruct the reading order. These systems rely on multiple task-specific models, including layout detection~\cite{rtdetr,layoutlmv3,yolov10}, formula detection~\cite{yolov8_ultralytics}, formula recognition~\cite{unimernet}, table recognition, text detection, text recognition~\cite{easyocr,ppocr3}, and reading order prediction~\cite{layoutreader}. Trained on massive datasets, general VLMs~\cite{got,monkey,textmonkey,internvl3.5,qwen2.5-vl, gemini3pro} have demonstrated strong potential for document parsing. End-to-end specialized VLMs such as olmocr~\cite{olmocr}, Nanonets-OCR~\cite{Nanonets-OCR-S}, and OCRFlux~\cite{ocrflux} further improve document parsing performance by fine-tuning the Qwen-VL~\cite{qwen2-vl, qwen2.5-vl, qwen3-vl} series on document parsing tasks. In addition, HunyuanOCR~\cite{hunyuanocr} and the dots.ocr series~\cite{dotsocr, dotsmocr} extend document parsing capabilities to support tasks such as visual question answering and SVG code generation. Meanwhile, DeepSeek-OCR 2~\cite{deepseek-ocr2} introduces a causal visual flow modeling mechanism to better capture the visual dependencies in document reading. Recent work~\cite{training-free} improves the inference efficiency of document parsing VLMs via training-free hierarchical speculative decoding. MonkeyOCR~\cite{monkeyocr} points out that traditional pipeline systems suffer from error accumulation due to the combination of multiple tools, while end-to-end models that directly process full-page documents may suffer from low efficiency and hallucination issues caused by long contexts. To address these limitations, it proposes a three-stage document parsing paradigm, SRR, consisting of structure detection, content recognition with a VLM, and relation prediction. Subsequently, PaddleOCR-VL~\cite{paddleocr-vl} adopts a similar three-stage framework. MinerU2.5~\cite{mineru2.5} and MonkeyOCR v1.5~\cite{monkeyocr1.5} further merge structure detection and relation prediction into a single vlm, simplifying the pipeline into a two-stage parsing framework.

\begin{figure}[tbp]
    \centering
    \includegraphics[width=0.9\textwidth]{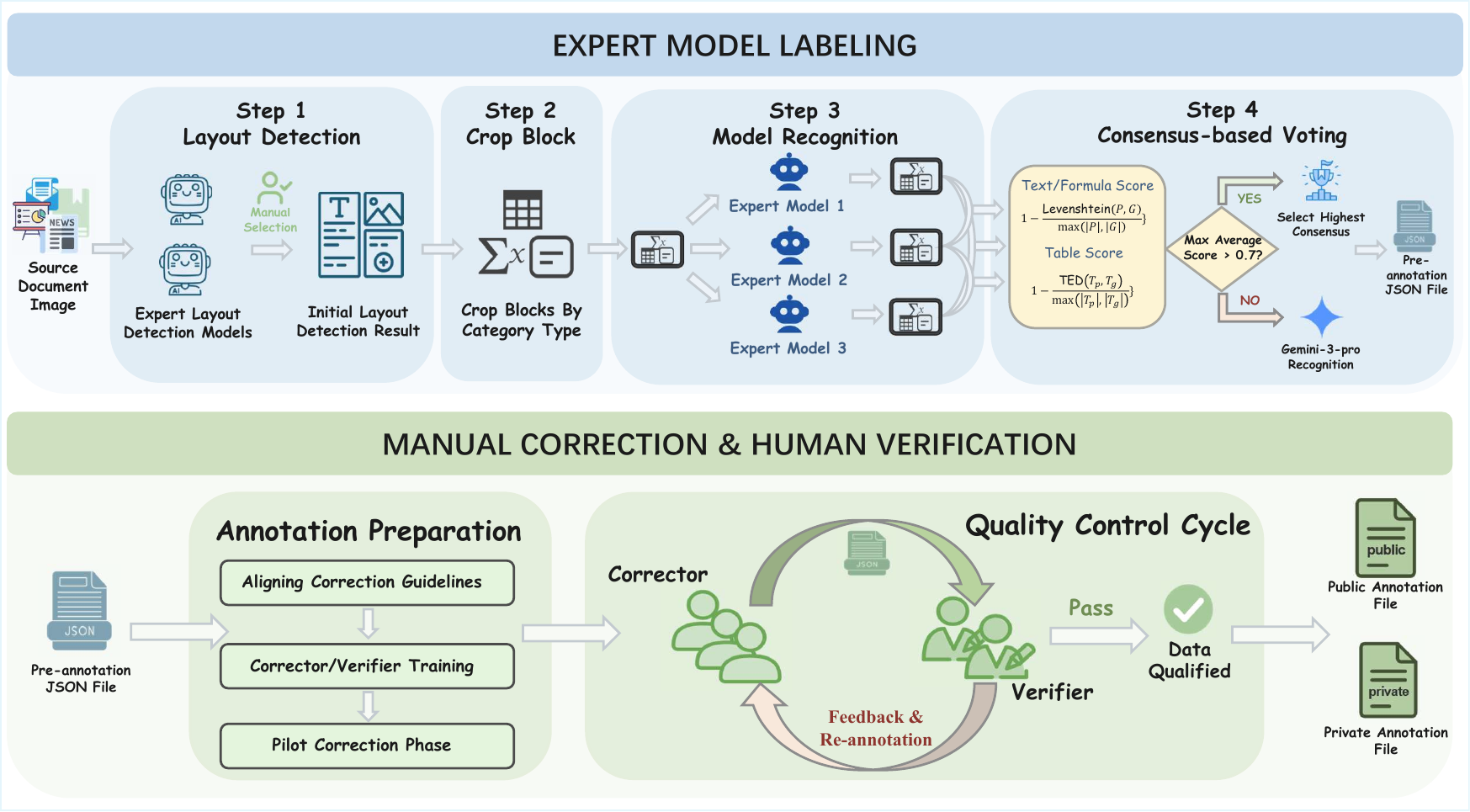}
    \caption{Overall pipeline of the data annotation process.}
    \label{fig:pipeline}
\end{figure}

\subsection{Document Parsing Benchmarks}

Early document parsing benchmarks predominantly focused on single, isolated tasks. For instance, M\textsuperscript{6}Doc~\cite{m6doc}, CDLA~\cite{cdla}, D4LA~\cite{d4la}, and DocLayNet~\cite{doclaynet} were designed for layout analysis; UniMER-1M~\cite{unimernet} and HME-100k~\cite{hme-100k} for formula recognition; and FinTabNet~\cite{fintabnet} and PubTabNet~\cite{pubtabnet} for table recognition. Recently, there has been a paradigm shift towards unified, multi-task evaluation frameworks. For example, FoxPage~\cite{foxpage} focuses on the structured domain of academic papers sourced from clean PDF documents. OCRBenchV2~\cite{ocrbenchv2} encompasses 23 tasks including document parsing to comprehensively assess document understanding capabilities. OmniDocBench~\cite{omnidocbench} introduces a multi-level document parsing evaluation framework covering full-page, module, and attribute levels across diverse document types, whereas olmOCR-Bench~\cite{olmocr} focuses on assessing content recall for document parsing models. However, these comprehensive benchmarks primarily focus on well-formatted, digitally born documents. To address real-world scenarios, recent efforts such as DocPTBench~\cite{docptbench} and Real5-OmniDocBench~\cite{real5-omnidocbench} have evaluated the parsing of captured documents by printing and photographing 1,355 document images from OmniDocBench v1.5. 
As shown in Tab.~\ref{tab:benchmark_final_pro}, existing open-source benchmarks remain confined to a limited set of languages, leaving a significant gap in multilingual document evaluation.

\section{MDPBench}
Constructing a benchmark for multilingual photographed document parsing is inherently challenging. The dataset must simultaneously ensure type diversity, annotation accuracy, and realistic visual conditions. We carefully design the construction pipeline of MDPBench with three key objectives: diversity, realism, and reliability. Fig.~\ref{fig:photo} presents a visualization of the dataset.
In the following subsections, we introduce the dataset construction process in detail, including data collection, annotation methodology, and evaluation metrics. Fig.~\ref{fig:pipeline} illustrates the overall pipeline of the data annotation process.

\subsection{Multilingual Digital-born Document Collection}
To ensure a comprehensive evaluation, we prioritize diversity in document types, layout complexity, and visual elements (e.g., formulas, images, tables, and charts) during data collection. We systematically source data from globally accessible public platforms, covering 17 representative languages. Our data sources include open-access academic papers, business reports, educational materials, handwritten notes, historical archives, modern newspapers, as well as challenging Chinese and English documents from OmniDocBench~\cite{omnidocbench}, and complex text-image documents such as textbooks and comics from public digital libraries. Following the collection phase, all samples undergo manual review to filter out low-quality or structurally trivial documents. Ultimately, we curate a dataset of 850 digital-born document images spanning 17 languages.

\begin{table}[tbp]
\centering
\caption{Performance of general VLMs, specialized VLMs, and pipeline tools on MDPBench.}
\label{tab:evaluation_results_combined_all}
\resizebox{\linewidth}{!}{
\begin{tabular}{l *{22}{c} | c}
\toprule
\multirow{2}{*}{\textbf{Model}} & \multicolumn{3}{c}{\textbf{Overall}} & \multicolumn{10}{c}{\textbf{Latin}} & \multicolumn{9}{c}{\textbf{Non-Latin}} & \multicolumn{1}{c}{\textbf{Private}} \\
\cmidrule(lr){2-4} \cmidrule(lr){5-14} \cmidrule(lr){15-23} \cmidrule(lr){24-24}
& All & Digit. & Photo. & Avg. & DE & EN & ES & FR & ID & IT & NL & PT & VI & Avg. & AR & HI & JP & KO & RU & TH & ZH & ZH-T & All \\
\midrule
\multicolumn{24}{l}{\textbf{\textit{General VLMs}}} \\
\midrule
Gemini-3-pro-preview~\cite{gemini3pro} & \textbf{86.4} & \underline{90.4} & \textbf{85.1} & \textbf{88.4} & \textbf{91.2} & \textbf{90.6} & \textbf{83.4} & \textbf{82.7} & \textbf{91.5} & \textbf{91.6} & \textbf{87.7} & \textbf{91.4} & \underline{85.9} & \textbf{84.1} & \textbf{89.4} & \textbf{90.4} & \underline{74.8} & \underline{85.5} & \textbf{84.9} & \textbf{80.6} & \textbf{85.1} & \textbf{82.1} & \textbf{89.8} \\
kimi-K2.5~\cite{kimi-k2.5} & 77.5 & 85.0 & 75.0 & 81.6 & \underline{85.9} & 86.2 & 72.7 & 71.0 & 80.6 & 86.6 & 77.4 & 87.6 & \textbf{86.2} & 72.9 & 75.8 & 74.5 & 72.5 & 70.9 & 61.8 & 67.0 & 81.7 & 78.6 & 81.2 \\
Doubao-2.0-pro~\cite{doubao} & 74.2 & 78.9 & 72.8 & 75.7 & 82.8 & 74.4 & 69.0 & 70.0 & 73.3 & 82.0 & 69.9 & 83.4 & 76.5 & 72.5 & 81.3 & 75.7 & 65.8 & 74.7 & 63.3 & 71.9 & 71.9 & 75.2 & 79.5 \\
Claude-Sonnet-4.6~\cite{claude} & 73.1 & 85.0 & 69.3 & 79.2 & 79.8 & 80.6 & 72.8 & 66.5 & 82.3 & 83.3 & 76.7 & 88.0 & 83.1 & 66.2 & 67.8 & 71.7 & 63.4 & 64.3 & 70.8 & 65.2 & 61.3 & 65.1 & 77.6 \\
ChatGPT-5.2-2025-12-11~\cite{chatgpt} & 68.6 & 85.6 & 63.0 & 75.2 & 70.8 & 79.4 & 71.4 & 60.0 & 77.7 & 78.5 & 71.6 & 85.0 & 82.1 & 61.1 & 64.9 & 63.4 & 55.8 & 65.4 & 60.7 & 63.8 & 56.3 & 58.7 & 74.0 \\
Qwen3-VL-Instruct-8b~\cite{qwen3-vl} & 68.3 & 78.4 & 65.0 & 73.6 & 73.7 & 71.4 & 69.3 & 66.2 & 68.5 & 79.1 & 78.3 & 82.2 & 73.4 & 62.5 & 63.1 & 58.4 & 59.9 & 61.9 & 57.9 & 62.0 & 62.6 & 73.8 & 70.8 \\
Qwen3.5-Instruct-9B~\cite{qwen3.5} & 65.7 & 74.8 & 62.7 & 72.5 & 72.8 & 72.0 & 72.0 & 64.4 & 66.2 & 77.6 & 74.5 & 79.1 & 74.0 & 58.2 & 53.4 & 56.2 & 55.7 & 60.3 & 54.7 & 56.7 & 60.8 & 67.5 & 68.9 \\
InternVL-3.5-8B~\cite{internvl3.5} & 42.7 & 59.7 & 37.0 & 53.4 & 39.8 & 64.2 & 47.5 & 42.7 & 53.8 & 60.6 & 52.2 & 63.2 & 57.0 & 30.6 & 8.2 & 9.0 & 45.6 & 30.3 & 26.1 & 10.8 & 55.3 & 59.3 & 45.3 \\
\midrule
\multicolumn{24}{l}{\textbf{\textit{Specialized VLMs}}} \\
\midrule
dots.mocr~\cite{dotsmocr} & \underline{80.5} & \textbf{90.5} & \underline{77.2} & \underline{81.7} & 82.6 & \underline{87.4} & 71.3 & 70.1 & \underline{84.5} & \underline{89.3} & \underline{83.2} & 86.8 & 79.9 & \underline{79.2} & \underline{83.3} & \underline{83.6} & \textbf{75.0} & 78.7 & 71.2 & \underline{77.9} & 84.6 & \underline{79.6} & \underline{82.8} \\
PaddleOCR-VL-1.5~\cite{paddleocr-vl-1.5} & 78.3 & 87.4 & 75.2 & 81.2 & 84.8 & 83.0 & 75.7 & \underline{78.1} & 83.9 & 85.2 & 80.6 & 80.2 & 78.9 & 74.9 & 71.3 & 67.7 & 69.5 & \textbf{86.0} & \underline{76.0} & 68.4 & \underline{84.8} & 75.7 & 80.7 \\
dots.ocr~\cite{dotsocr} & 76.5 & 88.8 & 72.3 & 79.1 & 79.7 & 81.2 & 69.2 & 67.1 & 82.5 & 87.8 & 78.8 & 86.9 & 79.1 & 73.5 & 75.9 & 77.3 & 70.6 & 68.5 & 66.8 & 73.3 & 79.1 & 76.2 & 79.7 \\
olmOCR2~\cite{olmocr2} & 70.4 & 79.9 & 67.2 & 76.7 & 75.7 & 77.3 & 72.5 & 68.9 & 70.6 & 81.0 & 72.0 & \underline{88.0} & 84.0 & 63.3 & 59.0 & 60.8 & 59.4 & 70.6 & 65.8 & 59.2 & 68.6 & 63.4 & 76.1 \\
PaddleOCR-VL~\cite{paddleocr-vl} & 69.6 & 87.6 & 63.6 & 72.1 & 78.2 & 79.3 & 62.9 & 66.0 & 77.4 & 78.4 & 67.9 & 72.0 & 66.6 & 66.7 & 65.8 & 68.4 & 59.9 & 77.8 & 56.9 & 57.8 & 78.2 & 68.5 & 70.9 \\
HunyuanOCR~\cite{hunyuanocr} & 68.3 & 80.2 & 64.3 & 72.4 & 75.0 & 73.1 & 63.0 & 66.1 & 69.9 & 80.3 & 61.4 & 81.9 & 80.6 & 63.7 & 68.3 & 73.1 & 55.6 & 68.9 & 52.2 & 60.7 & 66.8 & 64.2 & 68.6\\
GLM-OCR~\cite{glm-ocr} & 67.3 & 77.9 & 63.7 & 78.7 & 82.7 & 84.5 & \underline{75.8} & 76.2 & 79.7 & 82.8 & 80.2 & 77.4 & 69.2 & 54.3 & 21.7 & 39.6 & 65.5 & 61.2 & 64.2 & 27.4 & 78.5 & 76.7 & 68.8 \\
MonkeyOCRv1.5~\cite{monkeyocr1.5} & 65.0 & 84.3 & 58.6 & 67.4 & 70.8 & 74.9 & 55.6 & 60.3 & 73.8 & 75.9 & 66.3 & 67.2 & 61.4 & 62.4 & 60.1 & 56.8 & 57.0 & 78.9 & 51.7 & 55.6 & 74.8 & 64.1 & 69.0 \\
Nanonets-ocr2-3B~\cite{Nanonets-OCR2} & 64.2 & 79.2 & 59.3 & 71.4 & 76.7 & 76.4 & 61.8 & 66.1 & 68.4 & 78.5 & 74.1 & 74.2 & 66.0 & 56.2 & 60.2 & 59.2 & 52.1 & 54.7 & 45.5 & 44.6 & 68.3 & 65.1 & 67.6 \\
Nanonets-OCR-s~\cite{Nanonets-OCR-S} & 63.7 & 78.8 & 58.7 & 71.3 & 75.1 & 78.5 & 61.2 & 62.5 & 70.3 & 81.0 & 69.6 & 75.9 & 67.5 & 55.0 & 59.5 & 61.8 & 55.9 & 51.2 & 43.5 & 39.5 & 67.4 & 61.5 & 66.6 \\
MonkeyOCR-pro-3B~\cite{monkeyocr} & 52.2 & 68.0 & 47.0 & 65.1 & 71.7 & 77.9 & 55.9 & 62.1 & 66.2 & 74.5 & 66.3 & 71.1 & 40.2 & 37.6 & 4.6 & 4.2 & 55.2 & 60.5 & 42.6 & 9.1 & 72.2 & 52.4 & 53.6 \\
DeepSeek-OCR~\cite{deepseekocr} & 51.8 & 80.7 & 42.2 & 54.5 & 55.0 & 58.3 & 44.1 & 43.2 & 60.9 & 69.3 & 52.4 & 53.0 & 54.1 & 48.9 & 56.9 & 52.2 & 49.1 & 28.2 & 36.2 & 49.4 & 59.7 & 59.2 & 54.5 \\
MinerU-2.5-VLM~\cite{mineru2.5} & 46.3 & 61.9 & 40.8 & 63.0 & 68.8 & 78.4 & 54.7 & 57.3 & 67.5 & 75.2 & 60.4 & 58.8 & 46.0 & 27.4 & 1.3 & 9.0 & 39.1 & 14.7 & 8.6 & 11.3 & 72.9 & 62.2 & 48.7 \\
\midrule
\multicolumn{24}{l}{\textbf{\textit{Pipeline Tools}}} \\
\midrule
PP-StructureV3~\cite{ppstruct3} & 45.4 & 56.2 & 41.7 & 59.8 & 60.4 & 68.7 & 54.4 & 49.8 & 69.6 & 68.9 & 55.5 & 58.4 & 52.7 & 28.9 & 1.0 & 7.7 & 56.2 & 15.4 & 7.5 & 11.9 & 72.2 & 59.1 & 49.6 \\
MinerU-2.5-pipeline~\cite{mineru2.5} & 33.5 & 57.6 & 25.4 & 46.5 & 54.3 & 58.3 & 38.4 & 43.6 & 51.9 & 56.5 & 43.9 & 44.0 & 27.6 & 18.7 & 1.2 & 5.3 & 24.5 & 6.8 & 4.2 & 6.4 & 53.9 & 47.2 & 36.2 \\
\bottomrule
\end{tabular}
}
\end{table}

\subsection{Real-world Photographed Document Generation.}
To evaluate model robustness under real-world degradations, we transform the aforementioned digital-born documents into photographed data by printing them on paper or displaying them on computer screens for capture. 
We then photograph all documents in both indoor and outdoor environments. To simulate the complexities of real-world capture, we apply varying degrees of physical deformation to the printed documents, including inward bending, outward bending, and irregular wrinkling. In addition, we introduce diverse camera perspectives, capturing images from left, right, inverted, and oblique angles. For each document, we collect three images: two indoors and one outdoors.The indoor images feature diverse background interferences (e.g., desk surfaces, floor textures, and background text) and are subject to complex indoor lighting, moiré patterns (from screen captures), reflections, glare, and slight blur. Conversely, the outdoor images introduce distinct challenges, such as low-light conditions, shadows cast by surrounding objects, uneven illumination, and complex natural backgrounds.

\begin{figure}[tbp]
    \centering
    \includegraphics[width=1\textwidth]{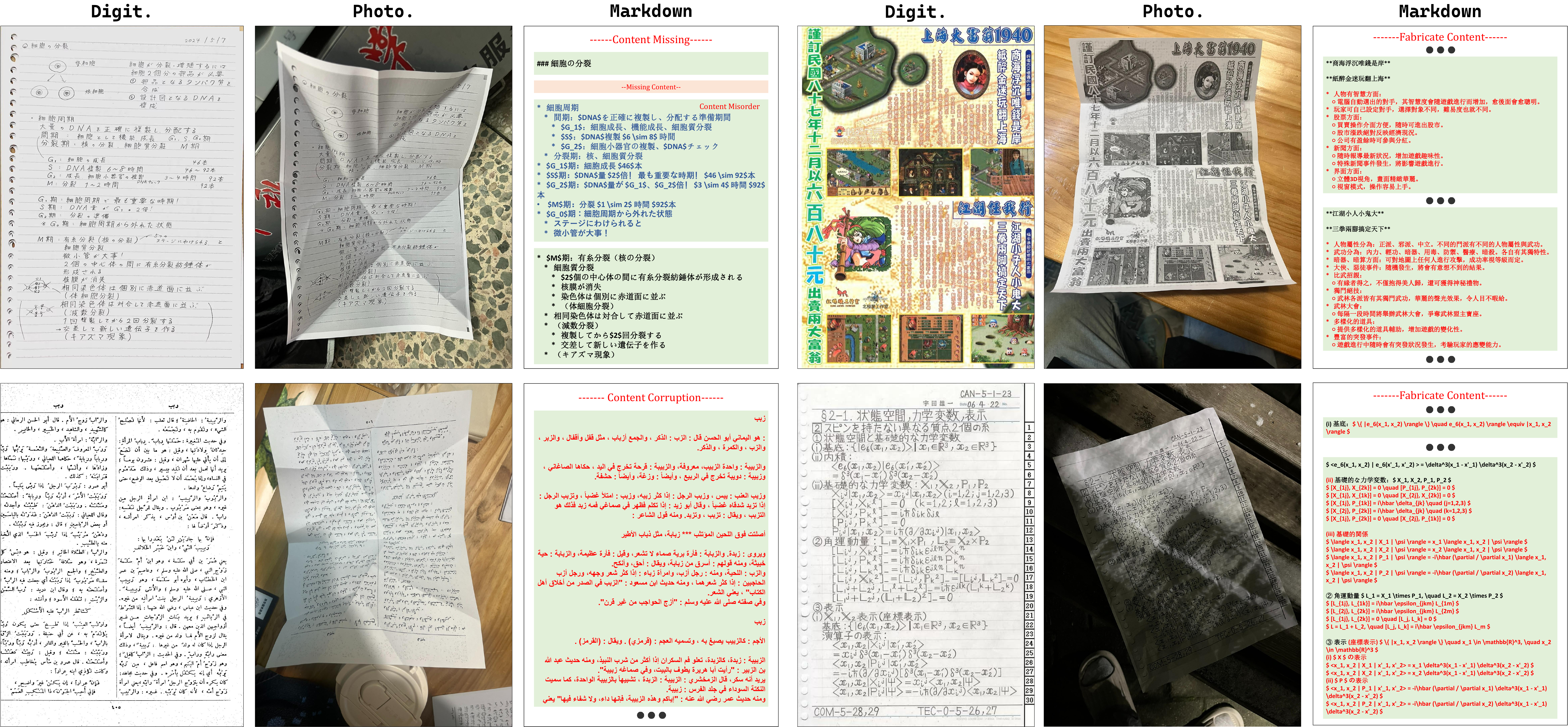}
    \caption{Visualization of document parsing results of Gemini-3-Pro on photographed documents.}
    \label{fig:MDPBench_gemini}
\end{figure}

\subsection{Data Annotation}
To balance efficiency and annotation accuracy, we design a three-stage annotation pipeline consisting of Expert Model Labeling, Manual Correction, and Human Verification. The resulting annotations follow a format that is fully compatible with OmniDocBench~\cite{omnidocbench}.

\textbf{Expert Model Labeling}. We first use dots.ocr~\cite{dotsocr} and PaddleOCR-VL~\cite{paddleocr-vl} to perform layout detection on all digital-born documents. The detection results from the two models are then manually compared for each image, and the one with fewer missed detections and false detections is selected as the initial layout annotation. Based on the obtained layout information, we crop the text blocks, table blocks, and formula blocks according to the bounding box coordinates and element types in the layout, producing individual element-level crops. We then employ three state-of-the-art recognition models, PaddleOCR-VL~\cite{paddleocr-vl}, dots.ocr~\cite{dotsocr}, and Qwen3VL~\cite{qwen3-vl}, to recognize these cropped elements. Since the correct recognition result is usually unique and stable, whereas incorrect results tend to be diverse and more random, the prediction that is most similar to the outputs of other models is more likely to be correct. Based on this observation, we compute the pairwise similarity among the recognition results of the three expert models for each element and select the result with the highest average similarity to the other two models as the final initial annotation. Specifically, for text and formulas, we measure similarity using 1 - Normalized Edit Distance (NED), while for tables, we use Tree Edit Distance-based Similarity (TEDS). If the highest average similarity is lower than 0.7, we consider the predictions of the three expert models to be unreliable. In such cases, we instead use the then state-of-the-art Gemini-3-pro~\cite{gemini3pro} model to recognize the corresponding element block.

\textbf{Manual Correction}. 
Before conducting manual corrections, we first train the annotators to align correction guidelines and introduce the annotation workflow. We then carry out a pilot annotation on a small subset of samples to verify and ensure the accuracy and consistency of the overall annotation process. After obtaining the model-generated annotations, we perform manual correction in a staged manner. First, we check whether the layout coordinates and element types of each image are correct. Next, we verify whether the reading order follows the natural reading logic of humans. Finally, we examine and refine each element detected in the layout one by one.

\textbf{Human Verification}. To ensure the final quality of the dataset, we adopt a strict ``annotation–correction–verification” mechanism. After manual correction of one document, the data is submitted to an independent reviewer for verification. If the annotation meets the quality standards, it is marked as ``Passed” and proceeds to the final delivery stage. If any errors or inconsistencies are identified, it is marked as ``Failed”, accompanied by detailed feedback, and returned to the original annotator for targeted revisions. This process is iteratively repeated until the document fully satisfies the acceptance criteria.

\begin{figure}[tbp]
    \centering
    \includegraphics[width=1\textwidth]{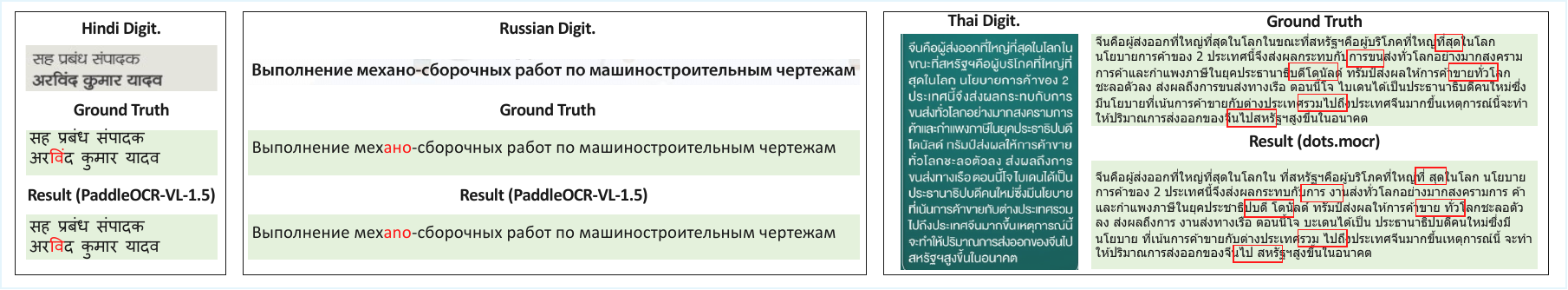}
    \caption{Typical language-specific errors.}
    \label{fig:lang_unique_mistake}
\end{figure}

\begin{figure}[tbp]
    \centering
    \includegraphics[width=1\textwidth]{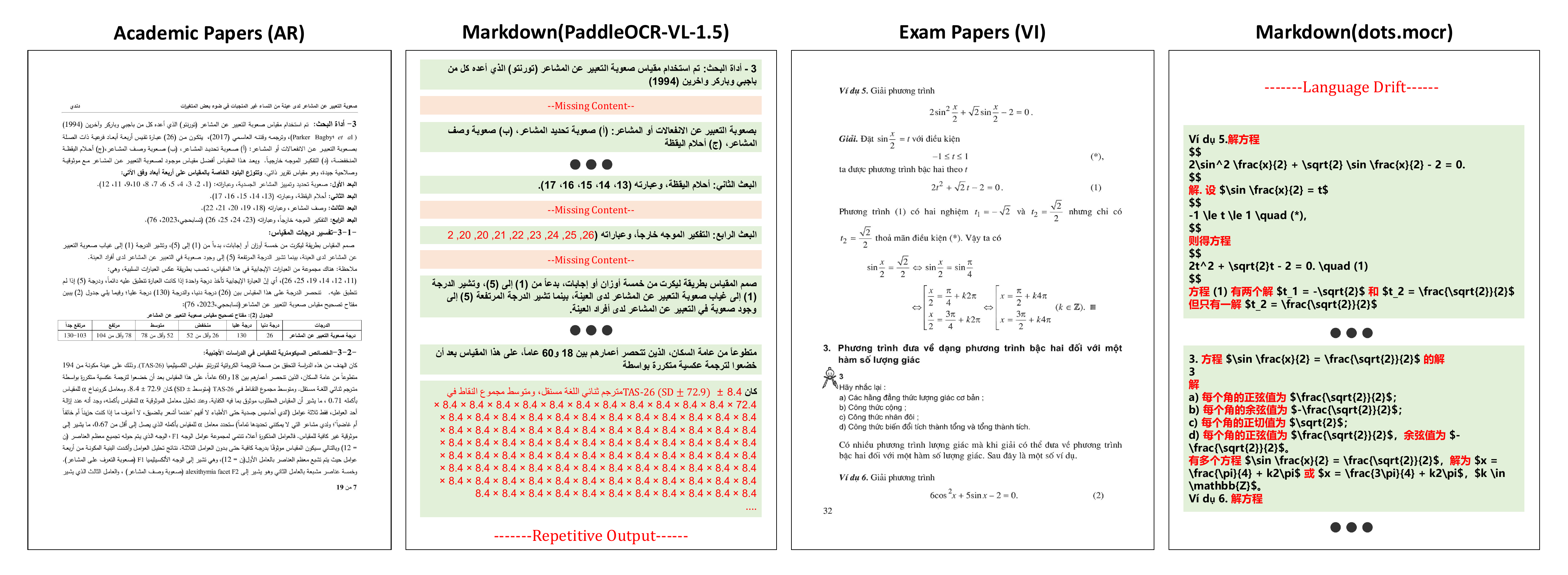}
    \caption{Existing document parsing models often exhibit hallucinations.}
    \label{fig:examples_illusion}
\end{figure}

\subsection{Evaluation Metrics}

Unlike OmniDocBench~\cite{omnidocbench}, MDPBench adopts a page-level aggregation evaluation strategy to mitigate the impact of imbalanced multilingual data distributions. In OmniDocBench~\cite{omnidocbench}, the overall metric is typically computed by first calculating the average scores of different element types—such as text, tables, and formulas—and then averaging these scores. However, in multilingual scenarios, document structures vary significantly across languages. For example, academic papers are often written in English and tend to contain a large number of formulas, while documents in some other languages may include far fewer formulas. If an element-level evaluation strategy is used in such cases, the overall score of certain languages can be disproportionately influenced by the parsing results of only a few formulas or tables. To address this issue, we adjust the evaluation granularity to the page level. Specifically, we first compute the evaluation metrics for all elements within a single page (e.g., text, tables, and formulas), and then average these metrics to obtain the score for that page. The final score is calculated as the average of the scores across all pages. To mitigate potential benchmark overfitting caused by targeted optimization on publicly available error cases, we divide the dataset into public and private subsets. The images and ground-truth annotations in the public subset will be released to the community for free download and evaluation. In contrast, the private subset and its annotations will remain undisclosed. Evaluation on the private subset can only be conducted by submitting model through our official evaluation website.

For each image, we follow OmniDocBench~\cite{omnidocbench} to perform data preprocessing, element extraction, pure text extraction, and element matching. During evaluation, page components such as headers, footers, page numbers, and page footnotes are ignored. For the matched elements, we use the following metrics for evaluation:
\begin{itemize}

    \item For text and reading order, we adopt Normalized Edit Distance (NED). Specifically, we compute the Levenshtein distance between the predicted text and the ground-truth text, and normalize it by the maximum length of the two strings:
    $$\text{Score} = 1 - \frac{\text{Levenshtein}(P, G)}{\max(|P|, |G|)}.$$

    \item For formula recognition, we use CDM~\cite{cdm} for evaluation to prevent misjudgments caused by differences in expression forms.

    \item For table recognition, we adopt the widely used Tree-Edit-Distance-based Similarity (TEDS)~\cite{pubtabnet}.

    $$\text{TEDS} = 1 - \frac{\text{TED}(T_p, T_g)}{\max(|T_p|, |T_g|)}$$
    
\end{itemize}

\section{Experiments}

We evaluate a diverse set of document parsing models on MDPBench, including general VLMs, specialized models, and pipeline-based tools. Beyond assessing document parsing performance, MDPBench also serves as a valuable benchmark for evaluating the multilingual text understanding capabilities of general VLMs. During evaluation, we assume that models have no prior knowledge of the input image’s language or whether it is photographed or digitally generated. The results are summarized in Tab.~\ref{tab:evaluation_results_combined_all}.

\begin{figure}[htbp]
    \centering
    \includegraphics[width=1\textwidth]{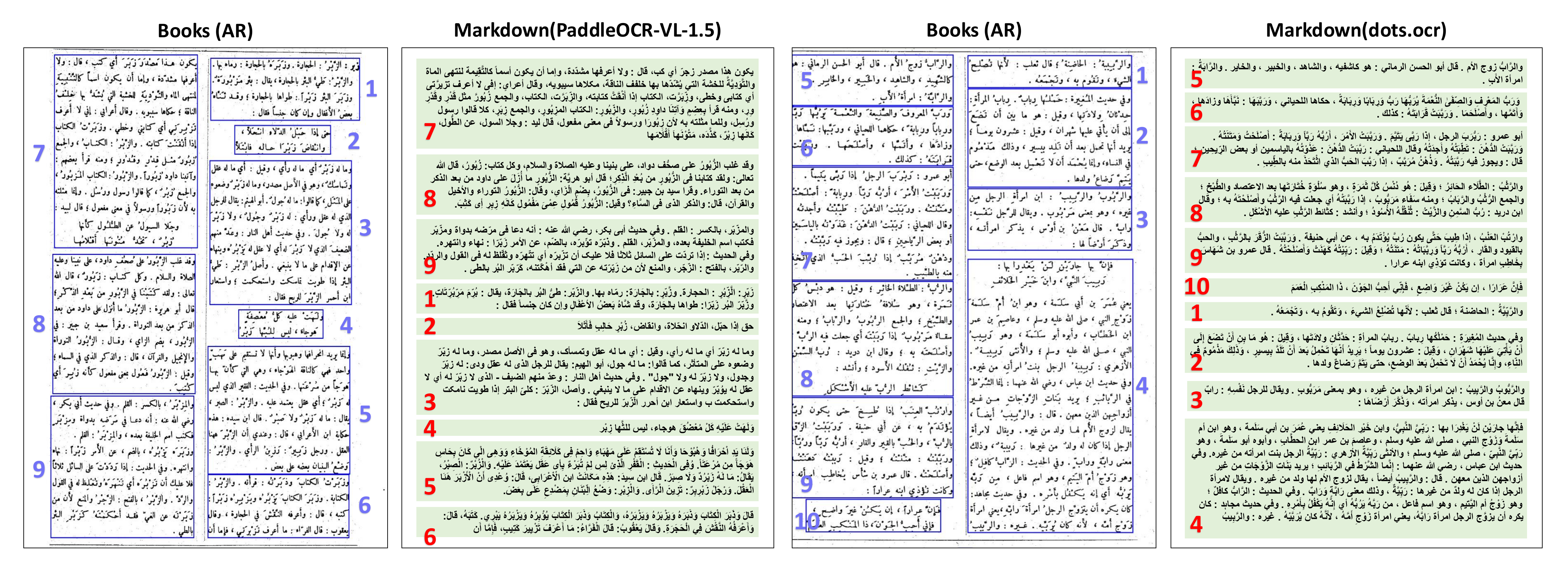}
    \caption{Existing document parsing models often fail to correctly handle right-to-left reading order in Arabic documents.}
    \label{fig:examples_wrong_order}
\end{figure}

\subsection{End-to-End Evaluation Results}

As shown in Tab.~\ref{tab:evaluation_results_combined_all}, the top-performing proprietary model, Gemini-3-Pro~\cite{gemini3pro}, achieves an overall accuracy of 86.4\%, reaching state-of-the-art (SOTA) results in 14 of 17 languages. In contrast, the best open-source model, dots.mocr, attains 80.5\% overall accuracy, revealing a clear gap between proprietary and open-source approaches. We further analyze the key limitations of current document parsing models, including challenges with photographed documents, limited recognition of non-Latin scripts, language-specific reading order issues, hallucinations and repeated outputs, and inherent drawbacks of traditional multi-stage pipelines. We summarize the findings as follows:

\textbf{Greater challenges in parsing photographed documents.} As shown in Tab.~\ref{tab:evaluation_results_combined_all}, model performance on photographed documents drops by an average of 17.8\%. Among all evaluated models, Gemini-3-Pro~\cite{gemini3pro} demonstrates the most robust performance on photographed documents. However, as illustrated in Fig.~\ref{fig:MDPBench_gemini}, even Gemini-3-Pro~\cite{gemini3pro} can still produce errors such as missing content, incorrect reading order, and hallucinated outputs when handling photographed documents, resulting in a performance decrease of approximately 5.3\% compared to digital documents.

\textbf{Models perform worse on non-Latin-script languages.} Latin-script languages are based on the A–Z alphabet, often extended with diacritics and other modifications. We observe that models perform significantly worse on non-Latin-script languages, with an average performance drop of 14.0\% compared to Latin-script languages. For example, although models such as MinerU2.5~\cite{mineru2.5} and MonkeyOCR~\cite{monkeyocr} are trained primarily on English and Chinese data, they still generalize well to Latin-script languages such as German. However, their accuracy drops to below 10\% on non-Latin-script languages, such as Arabic and Hindi.

% \textbf{Models exhibit typical errors specific to individual languages.} As illustrated in Fig.~\ref{fig:lang_unique_mistake}, models exhibit distinct language-specific error patterns when processing languages like Hindi, Russian, and Thai. In Hindi, which relies on vowel diacritics, models tend to retain only the base characters and ignore crucial modifiers, often misrecognizing `` {\devanagarifont अरविंद} " (Arvind) as `` {\devanagarifont अरविद} " (Aravid). For Russian documents, models frequently suffer from visual confusion, erroneously decoding Cyrillic characters that are visually identical to Latin letters (e.g., misclassifying the Cyrillic `` а ", `` н ", and `` о " in `` механо " as their Latin counterparts). Furthermore, when processing Thai, a typical unspaced language in which spaces denote only semantic boundaries, models often hallucinate spaces within continuous text. For instance, `` {\thaifont ที่ใหญ่ที่สุด} " (``biggest") is incorrectly segmented as `` {\thaifont ที่ใหญ่ที่ สุด} ", similar to splitting the English word ``biggest" into ``bigge st", thereby severely disrupting lexical integrity.

\textbf{Models exhibit typical errors specific to individual languages.} As illustrated in Fig.~\ref{fig:lang_unique_mistake}, models exhibit distinct language-specific error patterns when processing languages like Hindi, Russian, and Thai. In Hindi, which relies on vowel diacritics, models tend to retain only the base characters and ignore crucial modifiers, often misrecognizing ``\hindiArvind" (Arvind) as ``\hindiAravid" (Aravid). For Russian documents, models frequently suffer from visual confusion, erroneously decoding Cyrillic characters that are visually identical to Latin letters (e.g., misclassifying the Cyrillic ``\rusA", ``\rusH", and ``\rusO" in ``\rusAHO" as their Latin counterparts). Furthermore, when processing Thai, a typical unspaced language in which spaces denote only semantic boundaries, models often hallucinate spaces within continuous text. For instance, ``\thaiBiggest" (``biggest") is incorrectly segmented as ``\thaiBiggestWrong", similar to splitting the English word ``biggest" into ``bigge st", thereby severely disrupting lexical integrity.

\textbf{Models tend to produce repetitive outputs and exhibit language drift when handling multilingual documents.} As shown in Fig.~\ref{fig:examples_illusion}, we observe that PaddleOCR-VL-1.5~\cite{paddleocr-vl-1.5} exhibits issues such as missing content and repetitive outputs when processing Arabic document. Meanwhile, dots.mocr~\cite{dotsmocr} demonstrates language drift when handling Vietnamese, incorrectly recognizing it as Chinese. These findings suggest that existing document parsing models are insufficiently trained for such linguistic scenarios and suffer from biases in their training data.

\textbf{Models Struggle with Reading Order in Right-to-Left Scripts.} Arabic is read from right to left. As shown in Fig.~\ref{fig:examples_wrong_order}, we observe that for two-column Arabic document images, models such as PaddleOCR-VL-1.5~\cite{paddleocr-vl-1.5} and dots.ocr~\cite{dotsocr} often incorrectly process the text in a left-to-right order.

\begin{table}[tbp]
\centering
\caption{Component-level evaluation of text, formula, and table recognition on MDPBench subsets.}
\label{tab:merged_evaluation_results}
% Resize the table to fit the column/page width
\resizebox{\linewidth}{!}{
\begin{tabular}{l *{17}{c} | *{2}{c} | *{2}{c}}
\toprule
\multirow{2}{*}{\textbf{Model}} & \multicolumn{17}{c|}{\textbf{Text$^{\mathrm{Edit}}\downarrow$}} & \multicolumn{2}{c|}{\textbf{Formula$^{\mathrm{CDM}}\uparrow$}} & \multicolumn{2}{c}{\textbf{Table$^{\mathrm{TEDS}}\uparrow$}} \\
\cmidrule(lr){2-18} \cmidrule(lr){19-20} \cmidrule(lr){21-22}
& DE & EN & ES & FR & ID & IT & NL & PT & VI & AR & HI & JP & KO & RU & TH & ZH & ZH-T & Digit. & Photo. & Digit. & Photo. \\
\midrule
Gemini-3-pro-preview~\cite{gemini3pro} & 0.026 & 0.041 & \textbf{0.082} & 0.026 & 0.025 & 0.021 & 0.016 & 0.061 & 0.050 & \textbf{0.040} & \textbf{0.017} & 0.234 & 0.043 & 0.031 & \textbf{0.042} & 0.104 & 0.418 & \textbf{93.4} & \textbf{90.5} & 75.9 & \textbf{69.2} \\
Qwen3-VL-Instruct-8B~\cite{qwen3-vl} & 0.126 & 0.135 & 0.124 & \textbf{0.023} & 0.066 & 0.018 & 0.016 & 0.043 & 0.046 & 0.412 & 0.065 & 0.124 & 0.059 & 0.039 & 0.226 & 0.104 & 0.342 & 92.9 & 89.3 & 65.0 & 56.2 \\
dots.mocr~\cite{dotsmocr} & \textbf{0.003} & 0.022 & 0.107 & 0.027 & 0.021 & \textbf{0.011} & 0.005 & 0.024 & 0.039 & 0.059 & 0.024 & 0.071 & \textbf{0.031} & \textbf{0.030} & 0.598 & 0.084 & 0.804 & 89.8 & 78.7 & 59.6 & 55.9 \\
PaddleOCR-VL-1.5~\cite{paddleocr-vl-1.5} & \textbf{0.003} & \textbf{0.019} & 0.172 & 0.040 & \textbf{0.018} & 0.012 & \textbf{0.004} & \textbf{0.019} & \textbf{0.031} & 0.262 & 0.224 & \textbf{0.063} & \textbf{0.031} & 0.033 & 0.330 & \textbf{0.065} & \textbf{0.197} & 88.4 & 85.1 & \textbf{76.0} & 65.0 \\
MonkeyOCR-pro-3B~\cite{monkeyocr} & 0.016 & 0.027 & 0.169 & 0.035 & 0.461 & 0.016 & 0.018 & 0.040 & 0.913 & 0.978 & 0.980 & 0.139 & 0.113 & 0.458 & 0.989 & 0.126 & 0.660 & 90.7 & 88.3 & 68.3 & 60.7 \\
PP-StructureV3~\cite{ppstruct3} & 0.288 & 0.279 & 0.481 & 0.421 & 0.347 & 0.356 & 0.267 & 0.396 & 0.569 & 0.975 & 0.969 & 0.613 & 0.944 & 0.880 & 0.931 & 0.387 & 0.659 & 88.6 & 56.9 & 56.3 & 46.2 \\
\bottomrule
\end{tabular}
}
\end{table}

\subsection{Single Task Evaluation Results}

\textbf{Text Recognition Results.} We leverage the ground-truth layout annotations to crop text blocks from Digital-Born documents, and randomly sample 200 blocks per language for evaluation. The results, shown in Tab.~\ref{tab:merged_evaluation_results}, indicate that PaddleOCR-VL-1.5~\cite{paddleocr-vl-1.5} achieves the best performance on 10 out of 17 languages, while dots.mocr~\cite{dotsmocr} and Gemini-3-Pro attain state-of-the-art (SOTA) results on 4 languages each. This phenomenon reflects, to some extent, biases in both training data and training paradigms. Specifically, dots.mocr~\cite{dotsmocr} and Gemini-3-Pro~\cite{gemini3pro} are primarily trained in an end-to-end manner on full-page images, which leads to relatively weaker performance when handling cropped local text blocks. In contrast, PaddleOCR-VL-1.5~\cite{paddleocr-vl-1.5} is trained with a substantial amount of text-block-level data, making it better suited for this evaluation setting. Furthermore, we observe that PaddleOCR-VL-1.5~\cite{paddleocr-vl-1.5} performs notably worse on Arabic, Hindi, and Thai. This suggests a language distribution bias in its training data, which in turn limits its generalization ability on low-resource or underrepresented languages.

\noindent
\textbf{Formula and Table Recognition Results.} 
We crop formula and table regions from digital-born documents using ground-truth annotations, and manually extract corresponding regions from photographed documents. The results are reported in Tab.~\ref{tab:merged_evaluation_results}. Gemini-3-Pro~\cite{gemini3pro} achieves the best performance on both formula and table recognition. All models exhibit performance degradation in photographed scenarios, likely due to complex backgrounds, varying illumination, image degradation, and geometric distortions. Table recognition remains challenging: Gemini-3-Pro~\cite{gemini3pro} achieves 75.9\% accuracy on digital tables but drops to 69.2\% on photographed tables, indicating that table recognition still lacks robustness under real-world imaging conditions.

\noindent
\textbf{Layout Detection Results.} We evaluate different models on digital-born documents across multiple languages using the PageIoU~\cite{mineru2.5} metric. The results are shown in Tab.~\ref{tab:layout_detection_results}. dots.mocr~\cite{dotsmocr} achieves SOTA performance in 13 out of 17 languages, demonstrating strong generalization in multilingual scenarios. PaddleOCR-VL~\cite{paddleocr-vl} and PaddleOCR-VL-1.5~\cite{paddleocr-vl-1.5} exhibit comparable multilingual layout detection performance, resulting in similar overall results on digital-born documents. However, due to its support for arbitrarily shaped bounding boxes, PaddleOCR-VL-1.5~\cite{paddleocr-vl-1.5} improves overall performance on photographed documents by 11.6\%, as shown in Tab.~\ref{tab:evaluation_results_combined_all}.  Although MinerU-2.5-VLM achieves overall results below 10\% on AR, HI, and RU, its PageIoU scores on these three languages all exceed 85\%, indicating that layout detection performance is relatively insensitive to language differences.

\begin{table}[tbp]
\centering
\caption{Component-level layout detection evaluation on the MDPBench layout subset.}
\label{tab:layout_detection_results}
\resizebox{\linewidth}{!}{
\begin{tabular}{l *{17}{c}}
\toprule
\multirow{2}{*}{\textbf{Model}} & \multicolumn{17}{c}{\textbf{Layout Detection$^{\mathrm{PageIoU}}\uparrow$}} \\
\cmidrule(lr){2-18}
& DE & EN & ES & FR & ID & IT & NL & PT & VI & AR & HI & JP & KO & RU & TH & ZH & ZH-T \\
\midrule
dots.mocr~\cite{dotsmocr} & 93.1 & \textbf{86.6} & 88.3 & 85.9 & \textbf{94.4} & \textbf{92.3} & \textbf{93.6} & \textbf{92.2} & \textbf{92.6} & \textbf{95.4} & \textbf{92.7} & \textbf{91.6} & \textbf{91.6} & \textbf{92.9} & \textbf{90.2} & 85.7 & \textbf{89.9} \\
PaddleOCR-VL~\cite{paddleocr-vl} & \textbf{93.9} & 86.2 & \textbf{88.8} & \textbf{94.4} & 85.9 & 88.1 & 86.9 & 84.2 & 81.9 & 84.6 & 87.2 & 85.8 & 86.5 & 88.6 & 80.2 & \textbf{86.0} & 84.7 \\
PaddleOCR-VL-1.5~\cite{paddleocr-vl-1.5} & 92.4 & 84.5 & \textbf{88.8} & 93.5 & 84.3 & 88.0 & 86.2 & 81.3 & 80.6 & 87.5 & 86.6 & 86.5 & 86.8 & 87.2 & 81.0 & 84.9 & 84.6 \\
MinerU-2.5-VLM~\cite{mineru2.5} & 90.6 & 84.8 & 76.1 & 91.8 & 84.4 & 86.1 & 85.4 & 87.8 & 85.0 & 89.2 & 87.2 & 82.7 & 91.5 & 88.8 & 81.9 & 83.9 & 87.1 \\
PP-StructureV3~\cite{ppstruct3} & 65.5 & 63.7 & 58.8 & 70.9 & 61.0 & 62.1 & 62.0 & 59.0 & 56.0 & 59.3 & 66.5 & 59.7 & 62.5 & 65.8 & 55.7 & 59.8 & 58.7 \\
\bottomrule
\end{tabular}
}
\end{table}

\section{Conclusion}
This paper introduces MDPBench, the first benchmark for multilingual photographed document parsing. MDPBench comprises 3,400 high-quality human-annotated images covering 17 languages and systematically incorporates a wide range of real-world capture conditions. Extensive experiments demonstrate the limitations of existing document parsing models, particularly a significant performance degradation on non-Latin scripts and photographed document scenarios. MDPBench can be used not only to evaluate specialized document parsing systems but also as a benchmark for assessing the multilingual text understanding and OCR capabilities of general-purpose large multimodal models, providing insights for future model improvement and facilitating the development of more robust, generalizable, and practically deployable document parsing systems.

\bibliographystyle{splncs04}
\bibliography{main}

\begin{thebibliography}{10}
\providecommand{\url}[1]{\texttt{#1}}
\providecommand{\urlprefix}{URL }
\providecommand{\doi}[1]{https://doi.org/#1}

\bibitem{claude}
{Anthropic}: Claude. \url{https://www.anthropic.com/claude} (2025)

\bibitem{qwen3-vl}
Bai, S., Cai, Y., Chen, R., Chen, K., Chen, X., Cheng, Z., Deng, L., Ding, W., Gao, C., Ge, C., et~al.: Qwen3-vl technical report. arXiv preprint arXiv:2511.21631  (2025)

\bibitem{qwen2.5-vl}
Bai, S., Chen, K., Liu, X., Wang, J., Ge, W., Song, S., Dang, K., Wang, P., Wang, S., Tang, J., Zhong, H., Zhu, Y., Yang, M., Li, Z., Wan, J., Wang, P., Ding, W., Fu, Z., Xu, Y., Ye, J., Zhang, X., Xie, T., Cheng, Z., Zhang, H., Yang, Z., Xu, H., Lin, J.: Qwen2.5-vl technical report. arXiv preprint arXiv:2502.13923  (2025)

\bibitem{doubao}
{ByteDance}: Doubao. \url{https://research.doubao.com} (2026)

\bibitem{ocrflux}
ChatDoc: Ocrflux. \url{https://github.com/chatdoc-com/OCRFlux} (2025)

\bibitem{m6doc}
Cheng, H., Zhang, P., Wu, S., Zhang, J., Zhu, Q., Xie, Z., Li, J., Ding, K., Jin, L.: M6doc: a large-scale multi-format, multi-type, multi-layout, multi-language, multi-annotation category dataset for modern document layout analysis. In: Proceedings of the IEEE/CVF Conference on Computer Vision and Pattern Recognition. pp. 15138--15147 (2023)

\bibitem{paddleocr-vl}
Cui, C., Sun, T., Liang, S., Gao, T., Zhang, Z., Liu, J., Wang, X., Zhou, C., Liu, H., Lin, M., et~al.: Paddleocr-vl: Boosting multilingual document parsing via a 0.9 b ultra-compact vision-language model. arXiv preprint arXiv:2510.14528  (2025)

\bibitem{paddleocr-vl-1.5}
Cui, C., Sun, T., Liang, S., Gao, T., Zhang, Z., Liu, J., Wang, X., Zhou, C., Liu, H., Lin, M., et~al.: Paddleocr-vl-1.5: Towards a multi-task 0.9 b vlm for robust in-the-wild document parsing. arXiv preprint arXiv:2601.21957  (2026)

\bibitem{ppstruct3}
Cui, C., Sun, T., Lin, M., Gao, T., Zhang, Y., Liu, J., Wang, X., Zhang, Z., Zhou, C., Liu, H., et~al.: Paddleocr 3.0 technical report. arXiv preprint arXiv:2507.05595  (2025)

\bibitem{d4la}
Da, C., Luo, C., Zheng, Q., Yao, C.: Vision grid transformer for document layout analysis. In: Proceedings of the IEEE/CVF international conference on computer vision. pp. 19462--19472 (2023)

\bibitem{docptbench}
Du, Y., Chen, P., Ying, X., Chen, Z.: Docptbench: Benchmarking end-to-end photographed document parsing and translation. arXiv preprint arXiv:2511.18434  (2025)

\bibitem{glm-ocr}
Duan, S., Xue, Y., Wang, W., Su, Z., Liu, H., Yang, S., Gan, G., Wang, G., Wang, Z., Yan, S., et~al.: Glm-ocr technical report. arXiv preprint arXiv:2603.10910  (2026)

\bibitem{ocrbenchv2}
Fu, L., Kuang, Z., Song, J., Huang, M., Yang, B., Li, Y., Zhu, L., Luo, Q., Wang, X., Lu, H., Li, Z., Tang, G., Shan, B., Lin, C., Liu, Q., Wu, B., Feng, H., Liu, H., Huang, C., Tang, J., Chen, W., Jin, L., Liu, Y., Bai, X.: Ocrbench v2: An improved benchmark for evaluating large multimodal models on visual text localization and reasoning. In: Proceedings of the Neural Information Processing Systems Track on Datasets and Benchmarks (2025)

\bibitem{gemini3pro}
{Google DeepMind}: Gemini 3 pro. \url{https://blog.google/innovation-and-ai/technology/developers-tools/gemini-3-pro-vision} (2025)

\bibitem{layoutlmv3}
Huang, Y., Lv, T., Cui, L., Lu, Y., Wei, F.: Layoutlmv3: Pre-training for document ai with unified text and image masking. In: Proceedings of the 30th ACM international conference on multimedia. pp. 4083--4091 (2022)

\bibitem{easyocr}
{Jaided AI}: Easyocr: Ready-to-use ocr with 80+ supported languages. \url{https://github.com/JaidedAI/EasyOCR} (2024)

\bibitem{yolov8_ultralytics}
Jocher, G., Chaurasia, A., Qiu, J.: Ultralytics yolov8. \url{https://github.com/ultralytics/ultralytics} (2023)

\bibitem{ppocr3}
Li, C., Liu, W., Guo, R., Yin, X., Jiang, K., Du, Y., Du, Y., Zhu, L., Lai, B., Hu, X., et~al.: Pp-ocrv3: More attempts for the improvement of ultra lightweight ocr system. arXiv preprint arXiv:2206.03001  (2022)

\bibitem{cdla}
Li, H.: Cdla: A chinese document layout analysis (cdla) dataset. \url{https://github.com/buptlihang/CDLA} (2021)

\bibitem{dotsocr}
Li, Y., Yang, G., Liu, H., Wang, B., Zhang, C.: dots. ocr: Multilingual document layout parsing in a single vision-language model. arXiv preprint arXiv:2512.02498  (2025)

\bibitem{monkeyocr}
Li, Z., Liu, Y., Liu, Q., Ma, Z., Zhang, Z., Zhang, S., Guo, Z., Zhang, J., Wang, X., Bai, X.: Monkeyocr: Document parsing with a structure-recognition-relation triplet paradigm. arXiv preprint arXiv:2506.05218  (2025)

\bibitem{monkey}
Li, Z., Yang, B., Liu, Q., Ma, Z., Zhang, S., Yang, J., Sun, Y., Liu, Y., Bai, X.: Monkey: Image resolution and text label are important things for large multi-modal models. In: proceedings of the IEEE/CVF conference on computer vision and pattern recognition. pp. 26763--26773 (2024)

\bibitem{training-free}
Liao, W., Li, H., Xie, P., Cai, X., Shen, Y., Xin, Y., Qin, Q., Ye, S., Li, T., Hu, M., et~al.: Training-free acceleration for document parsing vision-language model with hierarchical speculative decoding. arXiv preprint arXiv:2602.12957  (2026)

\bibitem{foxpage}
Liu, C., Wei, H., Chen, J., Kong, L., Ge, Z., Zhu, Z., Zhao, L., Sun, J., Han, C., Zhang, X.: Focus anywhere for fine-grained multi-page document understanding. arXiv preprint arXiv:2405.14295  (2024)

\bibitem{textmonkey}
Liu, Y., Yang, B., Liu, Q., Li, Z., Ma, Z., Zhang, S., Bai, X.: Textmonkey: An ocr-free large multimodal model for understanding document. arXiv preprint arXiv:2403.04473  (2024)

\bibitem{docling}
Livathinos, N., Auer, C., Lysak, M., Nassar, A., Dolfi, M., Vagenas, P., Ramis, C.B., Omenetti, M., Dinkla, K., Kim, Y., et~al.: Docling: An efficient open-source toolkit for ai-driven document conversion. In: AAAI Conference on Artificial Intelligence (2025)

\bibitem{Nanonets-OCR-S}
Mandal, S., Talewar, A., Ahuja, P., Juvatkar, P.: Nanonets-ocr-s: A model for transforming documents into structured markdown with intelligent content recognition and semantic tagging (2025)

\bibitem{Nanonets-OCR2}
Mandal, S., Talewar, A., Thakuria, S., Ahuja, P., Juvatkar, P.: Nanonets-ocr2: A model for transforming documents into structured markdown with intelligent content recognition and semantic tagging (2025)

\bibitem{mineru2.5}
Niu, J., Liu, Z., Gu, Z., Wang, B., Ouyang, L., Zhao, Z., Chu, T., He, T., Wu, F., Zhang, Q., et~al.: Mineru2. 5: A decoupled vision-language model for efficient high-resolution document parsing. arXiv preprint arXiv:2509.22186  (2025)

\bibitem{chatgpt}
{OpenAI}: Chatgpt. \url{https://chat.openai.com} (2025)

\bibitem{omnidocbench}
Ouyang, L., Qu, Y., Zhou, H., Zhu, J., Zhang, R., Lin, Q., Wang, B., Zhao, Z., Jiang, M., Zhao, X., et~al.: Omnidocbench: Benchmarking diverse pdf document parsing with comprehensive annotations. In: Proceedings of the Computer Vision and Pattern Recognition Conference. pp. 24838--24848 (2025)

\bibitem{marker}
Paruchuri, V.: Marker. \url{https://github.com/datalab-to/marker} (2024)

\bibitem{doclaynet}
Pfitzmann, B., Auer, C., Dolfi, M., Nassar, A.S., Staar, P.: Doclaynet: A large human-annotated dataset for document-layout segmentation. In: Proceedings of the 28th ACM SIGKDD conference on knowledge discovery and data mining. pp. 3743--3751 (2022)

\bibitem{olmocr}
Poznanski, J., Borchardt, J., Dunkelberger, J., Huff, R., Lin, D., Rangapur, A., Wilhelm, C., Lo, K., Soldaini, L.: olmocr: Unlocking trillions of tokens in pdfs with vision language models. arXiv preprint arXiv:2502.18443  (2025)

\bibitem{olmocr2}
Poznanski, J., Soldaini, L., Lo, K.: olmocr 2: Unit test rewards for document ocr. arXiv preprint arXiv:2510.19817  (2025)

\bibitem{qwen3.5}
{Qwen Team}: {Qwen3.5}: Towards native multimodal agents (2026), \url{https://qwen.ai/blog?id=qwen3.5}

\bibitem{hunyuanocr}
Team, H.V., Lyu, P., Wan, X., Li, G., Peng, S., Wang, W., Wu, L., Shen, H., Zhou, Y., Tang, C., et~al.: Hunyuanocr technical report. arXiv preprint arXiv:2511.19575  (2025)

\bibitem{kimi-k2.5}
Team, K., Bai, T., Bai, Y., Bao, Y., Cai, S., Cao, Y., Charles, Y., Che, H., Chen, C., Chen, G., et~al.: Kimi k2. 5: Visual agentic intelligence. arXiv preprint arXiv:2602.02276  (2026)

\bibitem{yolov10}
Wang, A., Chen, H., Liu, L., Chen, K., Lin, Z., Han, J., et~al.: Yolov10: Real-time end-to-end object detection. Advances in Neural Information Processing Systems  \textbf{37},  107984--108011 (2024)

\bibitem{unimernet}
Wang, B., Gu, Z., Liang, G., Xu, C., Zhang, B., Shi, B., He, C.: Unimernet: A universal network for real-world mathematical expression recognition. arXiv preprint arXiv:2404.15254  (2024)

\bibitem{cdm}
Wang, B., Wu, F., Ouyang, L., Gu, Z., Zhang, R., Xia, R., Shi, B., Zhang, B., He, C.: Image over text: Transforming formula recognition evaluation with character detection matching. In: Proceedings of the Computer Vision and Pattern Recognition Conference. pp. 19681--19690 (2025)

\bibitem{mineru}
Wang, B., Xu, C., Zhao, X., Ouyang, L., Wu, F., Zhao, Z., Xu, R., Liu, K., Qu, Y., Shang, F., et~al.: Mineru: An open-source solution for precise document content extraction. arXiv preprint arXiv:2409.18839  (2024)

\bibitem{qwen2-vl}
Wang, P., Bai, S., Tan, S., Wang, S., Fan, Z., Bai, J., Chen, K., Liu, X., Wang, J., Ge, W., et~al.: Qwen2-vl: Enhancing vision-language model's perception of the world at any resolution. arXiv preprint arXiv:2409.12191  (2024)

\bibitem{internvl3.5}
Wang, W., Gao, Z., Gu, L., Pu, H., Cui, L., Wei, X., Liu, Z., Jing, L., Ye, S., Shao, J., et~al.: Internvl3. 5: Advancing open-source multimodal models in versatility, reasoning, and efficiency. arXiv preprint arXiv:2508.18265  (2025)

\bibitem{layoutreader}
Wang, Z., Xu, Y., Cui, L., Shang, J., Wei, F.: Layoutreader: Pre-training of text and layout for reading order detection. In: Proceedings of the 2021 Conference on Empirical Methods in Natural Language Processing. pp. 4735--4744 (2021)

\bibitem{got}
Wei, H., Liu, C., Chen, J., Wang, J., Kong, L., Xu, Y., Ge, Z., Zhao, L., Sun, J., Peng, Y., et~al.: General ocr theory: Towards ocr-2.0 via a unified end-to-end model. arXiv preprint arXiv:2409.01704  (2024)

\bibitem{deepseekocr}
Wei, H., Sun, Y., Li, Y.: Deepseek-ocr: Contexts optical compression. arXiv preprint arXiv:2510.18234  (2025)

\bibitem{deepseek-ocr2}
Wei, H., Sun, Y., Li, Y.: Deepseek-ocr 2: Visual causal flow. arXiv preprint arXiv:2601.20552  (2026)

\bibitem{hme-100k}
Yuan, Y., Liu, X., Dikubab, W., Liu, H., Ji, Z., Wu, Z., Bai, X.: Syntax-aware network for handwritten mathematical expression recognition. In: Proceedings of the IEEE/CVF conference on computer vision and pattern recognition. pp. 4553--4562 (2022)

\bibitem{monkeyocr1.5}
Zhang, J., Liu, Y., Wu, Z., Pang, G., Ye, Z., Zhong, Y., Ma, J., Wei, T., Xu, H., Chen, W., et~al.: Monkeyocr v1. 5 technical report: Unlocking robust document parsing for complex patterns. arXiv preprint arXiv:2511.10390  (2025)

\bibitem{rtdetr}
Zhao, Y., Lv, W., Xu, S., Wei, J., Wang, G., Dang, Q., Liu, Y., Chen, J.: Detrs beat yolos on real-time object detection. In: Proceedings of the IEEE/CVF conference on computer vision and pattern recognition. pp. 16965--16974 (2024)

\bibitem{dotsmocr}
Zheng, H., Li, Y., Zhang, K., Xin, L., Zhao, G., Liu, H., Chen, J., Lou, J., Qiu, J., Fu, Q., et~al.: Multimodal ocr: Parse anything from documents. arXiv preprint arXiv:2603.13032  (2026)

\bibitem{fintabnet}
Zheng, X., Burdick, D., Popa, L., Zhong, X., Wang, N.X.R.: Global table extractor (gte): A framework for joint table identification and cell structure recognition using visual context. In: Proceedings of the IEEE/CVF winter conference on applications of computer vision. pp. 697--706 (2021)

\bibitem{pubtabnet}
Zhong, X., ShafieiBavani, E., Jimeno~Yepes, A.: Image-based table recognition: data, model, and evaluation. In: European conference on computer vision. pp. 564--580. Springer (2020)

\bibitem{real5-omnidocbench}
Zhou, C., Gao, Z., Wang, X., Gao, T., Cui, C., Tang, J., Liu, Y.: Real5-omnidocbench: A full-scale physical reconstruction benchmark for robust document parsing in the wild. arXiv preprint arXiv:2603.04205  (2026)

\end{thebibliography}
\end{document}